\documentclass[10pt,twocolumn]{article} 
\usepackage{simpleConference}
\usepackage{times}
\usepackage{graphicx}
\usepackage{amssymb}
\usepackage{url,hyperref}
\usepackage{subcaption}
\usepackage{caption}
\usepackage{longtable}

\usepackage{booktabs, multirow} 
\usepackage{soul}
\usepackage[table]{xcolor} 
\usepackage{changepage,threeparttable} 

\newcommand{\source}[1]{\caption*{Source: {#1}} }

\graphicspath{ {./images/} }

\begin{document}

\title{A naive method to discover directions in the StyleGAN2 latent space}

\author{
\\
  Giardina, Andrea\\
  \texttt{andrea.giardina@open.ac.uk}
  \and
  \\
  Soumya Subhra Paria\\
  \texttt{soumya.paria@open.ac.uk}
  \and
  \\
  Adhikari, Kaustubh \\
  \texttt{kaustubh.adhikari@open.ac.uk}
  \\
  \\
  \\
  \\
}
\maketitle
\thispagestyle{empty}

\begin{abstract}
Several research groups have shown that Generative Adversarial Networks (GANs) can
generate photo-realistic images in recent years. Using the GANs, a map is created between a
latent code and a photo-realistic image. This process can also be reversed: given a photo as
input, it is possible to obtain the corresponding latent code. In this paper, we will show how
the inversion process can be easily exploited to interpret the latent space and control the
output of StyleGAN2, a GAN architecture capable of generating photo-realistic faces. From a
biological perspective, facial features such as nose size depend on important genetic factors,
and we explore the latent spaces that correspond to such biological features, including
masculinity and eye colour. We show the results obtained by applying the proposed method
to a set of photos extracted from the CelebA-HQ database. We quantify some of these
measures by utilizing two landmarking protocols, and evaluate their robustness through
statistical analysis. Finally we correlate these measures with the input parameters used to
perturb the latent spaces along those interpretable directions. Our results contribute towards
building the groundwork of using such GAN architecture in forensics to generate photo-realistic faces that satisfy certain biological attributes.
\end{abstract}

\section{Introduction}
With the rise of in-depth characterization of the face through landmarks \cite{PMID:29343858}, \cite{https://doi.org/10.1002/humu.22054}, assessing various characteristics of the
faces, such as measurements and shapes, as well as modifying those characteristics, such as
those used in real-time social media apps, have become increasingly common. These
characteristics, in particular quantitative representations such as measurements and shapes
have also been used to perform genetic analyses to elucidate genes associated with cranio-
facial morphology. Other characteristics, such as facial symmetry, masculinity, and
attractiveness have also been assessed, either algorithmically or through surveys \cite{https://doi.org/10.1002/ajpa.22688},
\cite{sym12020236},\cite{ekrami}. \\
Another interesting problem has been to reverse this, by using genetic information to predict
a person’s facial appearance
\cite{claes},
\cite{6d2323d09ccf4284a96200b268cad0a5},
\cite{CLAES2014208}. This has widespread potential
applications, from forensics to IVF screening. This problem can be decomposed into two
parts: one, to utilize the existing knowledge about genetic contributions to various facial
characteristics by predicting the holistic appearance; second: to produce photo-realistic faces
that properly represent the predicted appearance. For example, genes such as PAX1, PAX3,
DCHS2, RUNX2, EDAR etc. have been identified that have distinct contributions to different
parts of the face \cite{e9b2de052ba24465acb7b2a25134ee66}
 – some shape the soft tissues of the
face such as cartilages, and some shape the cranial bones underneath . Some genes impact the
width of the nose, another may affect its protrusion. Based on summary statistics from
existing genome-wide association studies (GWAS) published from our CANDELA cohort
and other cohorts \cite{10.1038/ncomms11616}, \cite{doi:10.1126/sciadv.abc6160}, \cite{10.7554/eLife.49898}, it is possible to build prediction models for each of these specific
characteristics. However, each of those predictions about an individual aspect of the face
need to be combined together to make a consistent prediction of the overall face morphology.
In addition, it is preferable to implement a procedure which generate photorealistic faces, and
which contains the capability to produce multiple faces with the same key characteristics to
give a range of variability, including but not limited to variations on accessories such as facial
hair or glasses.

To this objective, we tested the underlying architecture of the StyleGAN2 deep neural
network by exploring its underlying parameter space, linking the parameter space to
interpretable facial characteristics, and by generating faces with perturbations of those
parameters in the directions of certain characteristics.

\section{Background}
\begin{figure}
  \centering
         \includegraphics[width=5cm]{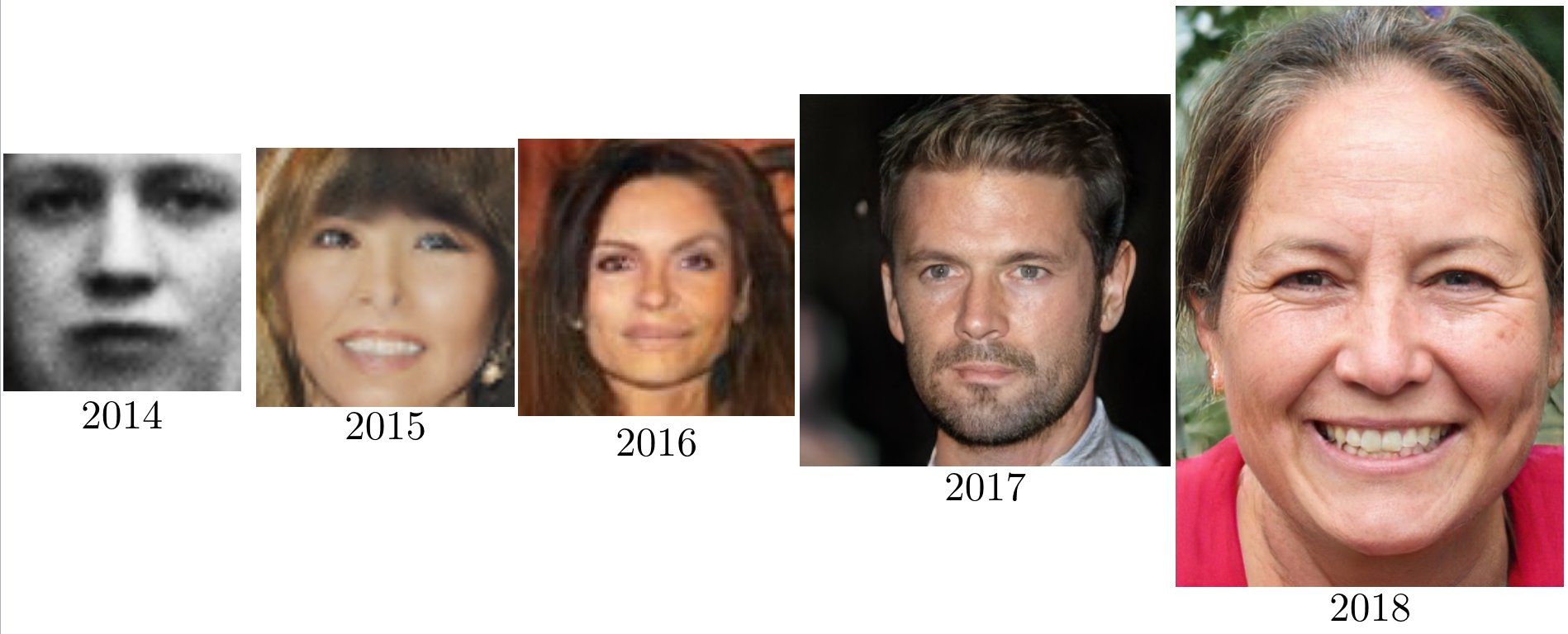}
  \caption{GAN progress on face generation.}
  \label{fig:ganprog}
  \source{\url{https://twitter.com/goodfellow_ian/status/1084973596236144640}}
\end{figure}

The Generative Adversarial Network (GAN) is a generative framework that works thanks to two antagonistic models.
A \(G\) model, called the Generator, tries to create synthetic data similar to real data, while a \(D\) model, called the Discriminator, tries to discriminate between synthetic and real data. The two models improve each other until the Discriminator can no longer distinguish between real and synthetic data. \\
GANs were introduced in a seminal paper by Ian Goodfellow \textit{et al.} in 2014  \cite{goodfellow2014generative}. Since then, the ability to generate photorealistic images has drastically improved, as exemplified beautifully by Goodfellow himself in an image posted on Twitter showing the progress made in the automatic generation of faces between 2014 and 2018  [Figure \ref{fig:ganprog}]. \\

Several studies have contributed to improving the performance of the gans, see among others \cite{radford2016unsupervised} and \cite{liu2016coupled}. A study on which to dwell is  ProGAN, introduced by Tero Karras \textit{et al.} in the Paper  \textit{Progressive Growing of Gans for Improved Quality, Stability, And Variation}. \\
Before the introduction of ProGAN, the generation of high-resolution images was a challenging task, given that the high resolution of the images made it extremely simple for the discriminator to separate synthetic images from real images \cite{odena2017conditional}. \\
The innovation at the base of ProGAN was to increase both the generator and discriminator progressively. The training process starts with low-resolution images of 4x4 pixels. Progressively the resolution of the images is doubled, and new layers are added simultaneously to the generator and discriminator, increasing the spatial resolution. One of the advantages of this type of architecture, which will then be the basis of future innovations \cite{karras2019stylebased}, \cite{karras2020analyzing}, is that the neural network initially discovers the basic structure of the images and progressively moves its attention to the finer details. \\
The same working group that participated in the realization of ProGAN has given life to StyleGAN\cite{karras2019stylebased} and the subsequent StyleGAN2\cite{karras2020analyzing} improved implementation. The main feature of Stylegan, which distinguishes this architecture from previous jobs, is the ability to control the style of the generated images in a certain way. The architecture allows the automatic separation of high-level attributes, such as pose or genre, from stochastic variations, such as the position of individual tufts of hair. \\
The GAN generator function maps typically the values sampled by a normal distribution to images . These codes are often called latent codes and the latent space where these vectors belong \(Z\) \cite{xia2021gan}.\\
StyleGAN does not map the latent code \(z\) directly to an image, but it maps it first to an intermediate latent vector \(w\) in the latent space \(W\). In the case of standard resolution at \(1024^2\), the function f that maps \(Z\) to \(W\) is a neural network at 8 layers, while the generator function \(g\) is a neural network at 18 layers. Both \(z\) in \(Z\) and \(w\) in \(W\)  are 512 dimension vectors. The innovation introduced into ProGAN, i.e. the training for subsequent resolutions, is also exploited in StyleGAN. At each resolution, starting at \(4^2\) up to \(1024^2\), two layers are associated. The previously calculated \(w\) vector feeds each generator layer. Vector \(w\) controls the style of the image. Depending on the resolution which is applied, only certain aspects of the images are affected: to the lowest resolutions (\(4^2-8^2\)) high-level aspects, such as the installation and shape of the face, to intermediate resolutions (\(16^2-32^2\)) more minor aspects like hairstyle, eyes closed / open, while at more elevating resolutions (\(64^2-1024^2\)) the finest aspects such as face microstructures. When the vector \(w\) is different for each layer, the relative latent space is called \(W^+\). If a vector \(w \in W\)  has size 512, a vector \(w \in W^+\) will have size 18x512.

\section{GAN Inversion}
Gan Inversion is the process that, given a photo, reconstructs the latent code in a pre-trained GAN model. The methods used can be grouped into three types: optimization-based, learning-based and hybrid. \\
Optimization-based methods try to optimize the latent vector for each image to minimize the difference with the original image at the level of pixels. \\
The methods based on learning, contrast, train an encoder network to map an image in a latent vector of the selected GAN  \cite{richardson2021encoding}, \cite{tov2021designing}. \\
The third approach, the hybrid one, tries to take the best of both methods. First, a latent vector is calculated via an encoder, and then the calculated latent vector is used to initialize the optimization network \cite{xia2021gan}. \\
Optimization-based methods achieve the best reconstruction results, but it is evident that performing an image optimization per image requires more execution time and more computational resources. Another disadvantage of optimization-based methods is the lower editability of the latent space compared to learning-based methods \cite{tov2021designing}. For editability, we mean the possibility of modifying the latent space to modify the given image meaningful and realistic.

\subsection{Restyle and pSp}
Among many GAN inversion methods, we have adopted ReStyle \cite{alaluf2021restyle}. \\
ReStyle is a learning-based method that uses an encoder for reversing an image. ReStyle, compared to the typical encoder-based inversion methods, does not infer an image into a single step but uses an iterative refinement process. At each step, the encoder is fed by the output of the previous step. This technique, called iterative error feedback or IEF, was introduced by Carreira \textit{et al.} \cite{carreira2016human} to estimate human pose.\\
The iterative scheme used by ReStyle is independent of the inversion function adopted and can therefore be applied to various encoders. The encoder used in our experiments is Pixel2Style2Pixel (pSp) \cite{richardson2021encoding}. pSp is a recent encoder introduced in 2021 whose results are at state of the art but manages to produce even better results once combined with the ReStyle iterative process.

\section{Latent space manipulation}
The generator in a GAN model can be seen as a function that maps a specific latent space \(Z\) to an image space \(X\). All images within \(X\) share the same semantic attributes. If, for example, X is the space of the facep generated by StyleGAN, every \(x \in X\) can be classified for various attributes, such as eye colour, face shape, hairstyle, age and gender.
The inversion of an image into a latent code is functional to the semantic manipulation of the image itself. For semantic manipulation, we mean the ability to modify the image, not going to act directly on the pixels of the image, as in the case of a classic photo-editing program, but instead on the high-level semantic attributes above. \\
As demonstrated by the paper InterfaceGAN\cite{shen2020interfacegan}, for each binary attribute, such as male/female, a hyperplane separates the latent codes belonging to one group rather than the other. Since two similar latent codes generate similar images, the normal vector to the hyperplane identifies a direction to change the attribute gradually from one group to the other. For example, say one hyperplane separates the latent codes of female subjects from men, the normal vector to this hyperplane in one direction will make the subject more masculine, while in the opposite direction more feminine. \\
An open search field is research for meaningful directions within the latent space. To find directions, InterfaceGAN has first summarized 500k images. Then, a pre-trained neural network hasc assigned a numeric score to each image related to the attribute of interest, such as glasses/no glasses. The InterfaceGAN authors then selected 10k images with the highest score and 10k with the lowest score and used these to train a linear SVM and thus discover the relative hyperplane that separates images belonging to the two groups. \\
The method adopted by InterFacegan is undoubtedly effective in finding any direction. However, it requires a heavy training phase and, above all, presupposes the existence of a pre-trained classifier for the attribute of interest (or that a classifier is trained ad hoc ). \\
For specific directions, particularly those concerning the somatic traits of a face, a naive method could help discover the directions.

\section{Our method}
The primary purpose of our study is to manipulate the somatic features of a face in the photo with a good level of precision. \\
Let \(g\) be a generating function that given a latent code \(z \in Z\) can obtain an image \(y \in Y\). Let \(p\) be a projection function that given an image \(x \in X \supset Y\) can obtain a latent code in \(Z\) such that \(g(p(x)) \sim x\). Let \(x_a\) and \(x_b\) in \(X\) be two identical images of the same subject but for the somatic characteristic \(s\) that we are interested in manipulating and \(z_a = p (x_a)\) and \(z_b = p (x_b\)) the respective latent vectors. \\
As amply demonstrated\cite{karras2019stylebased}, the corresponding synthesized images change linearly by linearly interpolating two latent codes. It follows that given \(v = z_b - z_a\) and \(g (z_a + \alpha * v)\) where \(\alpha \in [0,1]\) by changing the \(\alpha\) value we will be able to control the somatic trait \(s\), to bring it from the state \(a\), corresponding to image \(x_a\) at state \(b\) corresponding to image \(x_b\). \\
It is natural to wonder if the vector \(v\) is functional only to interpolate the photo \(x_a\) and \(x_b\) or if it can generalize on any photo. In other words, if photo \(x_a\) showed, for example, a subject with brown eyes and photo \(x_b\) the same subject with blue eyes, vector \(v\) would be able to transform the photo of any subject with brown eyes into eyes blues?  We have tried to answer this question in the experiments we have carried out, but it is first essential to consider the calculated vectors. \\ \\

\begin{figure*}
     \centering
     \begin{subfigure}[b]{0.3\textwidth}
         \centering
         \includegraphics[width=\textwidth]{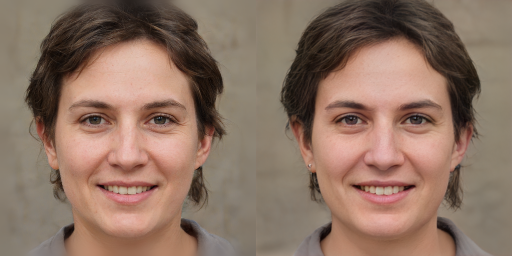}
         \caption{Subject 1}
         \label{fig:subject1}
     \end{subfigure}
     \hfill
     \begin{subfigure}[b]{0.3\textwidth}
         \centering
         \includegraphics[width=\textwidth]{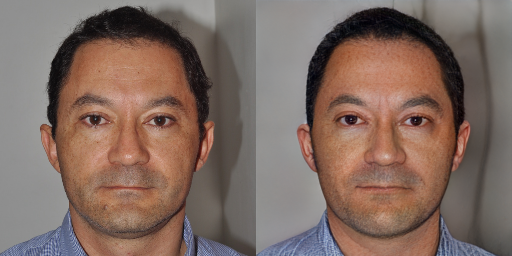}
         \caption{Subject 2}
         \label{fig:subject2}
     \end{subfigure}
     \hfill
     \begin{subfigure}[b]{0.3\textwidth}
         \centering
         \includegraphics[width=\textwidth]{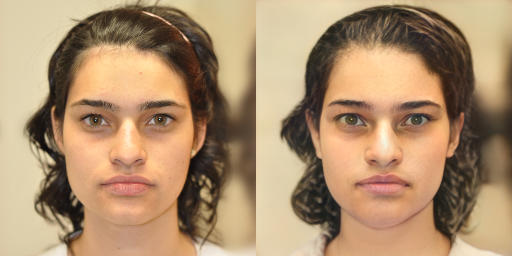}
         \caption{Subject 3}
         \label{fig:subject3}
     \end{subfigure}
        \caption{On the left side is the original photo, and on the right one, the image generated using the latent code calculated by ReStyle}
        \label{fig:subjects}
\end{figure*}

Figure \ref{fig:subjects} shows the subjects that were used as the basis for the calculation of the vectors. For each subject on the left we display the original photo, while on the right the synthesized photo using the latent vector calculated by ReStyle. Note that subject 1, unlike subjects 2 and 3, is not real but synthetic, generated by StyleGAN2. As expected, the projection of subject 1 is much closer to the original than the projections of subjects 2 and 3.

\begin{figure*}
     \centering
     \begin{subfigure}[b]{0.19\textwidth}
         \centering
         \includegraphics[width=\textwidth]{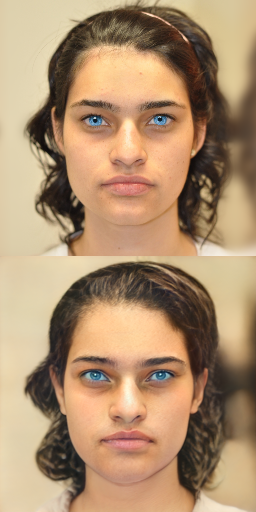}
         \caption{Eyes colour}
         \label{fig:eyescolour}
     \end{subfigure}
     \hfill
     \begin{subfigure}[b]{0.19\textwidth}
         \centering
         \includegraphics[width=\textwidth]{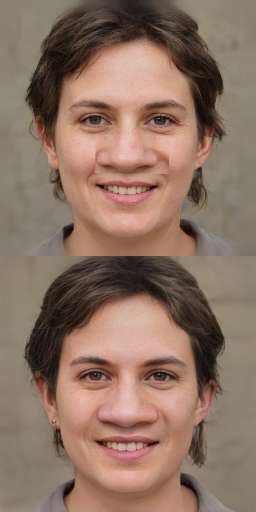}
         \caption{Nose}
         \label{fig:nose}
     \end{subfigure}
     \hfill
     \begin{subfigure}[b]{0.19\textwidth}
         \centering
         \includegraphics[width=\textwidth]{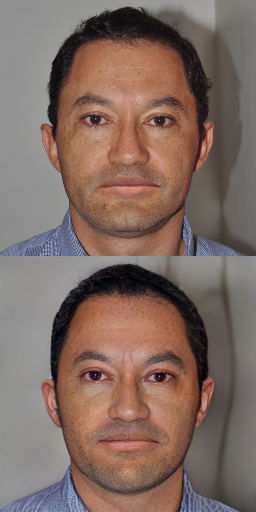}
         \caption{Lips}
         \label{fig:lips}
       \end{subfigure}
     \hfill
     \begin{subfigure}[b]{0.19\textwidth}
         \centering
         \includegraphics[width=\textwidth]{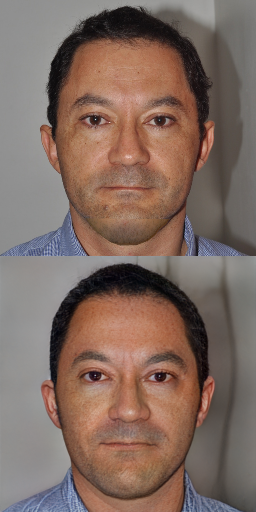}
         \caption{Chin}
         \label{fig:chin}
       \end{subfigure}
     \hfill
     \begin{subfigure}[b]{0.19\textwidth}
         \centering
         \includegraphics[width=\textwidth]{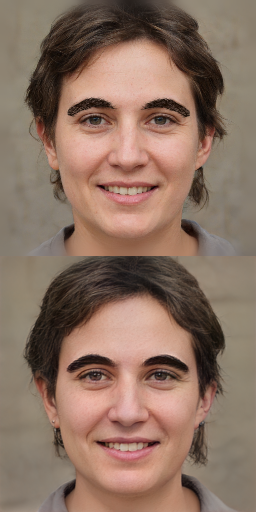}
         \caption{Eyebrow}
         \label{fig:eyebrow}
       \end{subfigure}
        \caption{The images modified with the photo editing program are visible in the upper line, while the images generated from the latent code calculated by ReStyle are shown in the lower line.}
        \label{fig:modified}
\end{figure*}

We looked at five facial features: eye colour, chin size, lip thickness, eyebrow thickness and nose size. For each of these somatic traits, we have chosen a starting photo, and through a photo-editing program, we have modified the same photo to change the somatic trait of interest. \\

Figure \ref{fig:modified} shows the five modified images for the somatic trait of interest. The images modified with the photo editing program are visible in the upper line, while the images generated from the latent code calculated by ReStyle are shown in the lower line. Note how the photos were edited in a very rough way, but despite this, ReStyle was able to grasp the significant aspects. In particular, the nose starting image is broken, but the projector has generated a realistic, albeit slightly caricatured image. 
The ability to reconstruct broken photos is an aspect to consider if a more accurate projector function is taken into account for future works: a more accurate projection function, perhaps based on an optimization process and not an encoding process, could have difficulty creating image projections of poor quality and not photorealistic sources, and therefore inadequate to work with this type of images. \\

\begin{figure*}
  \captionsetup[subfigure]{labelformat=empty}
  \centering
  \begin{subfigure}[b]{1\textwidth}
    \centering
         \includegraphics[width=\textwidth]{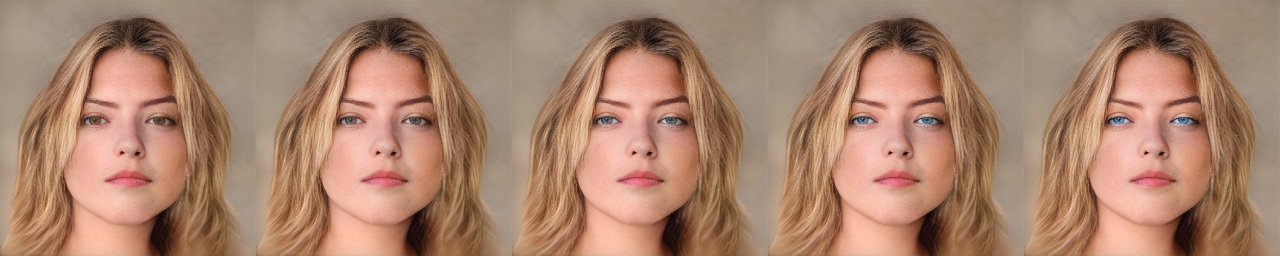}
         \caption{Less blue \(\longleftarrow\) Eyes colour \(\longrightarrow\) More blue}         
         \label{fig:eyescolour}
     \end{subfigure}
     \hfill
     \begin{subfigure}[b]{1\textwidth}
         \centering
         \includegraphics[width=\textwidth]{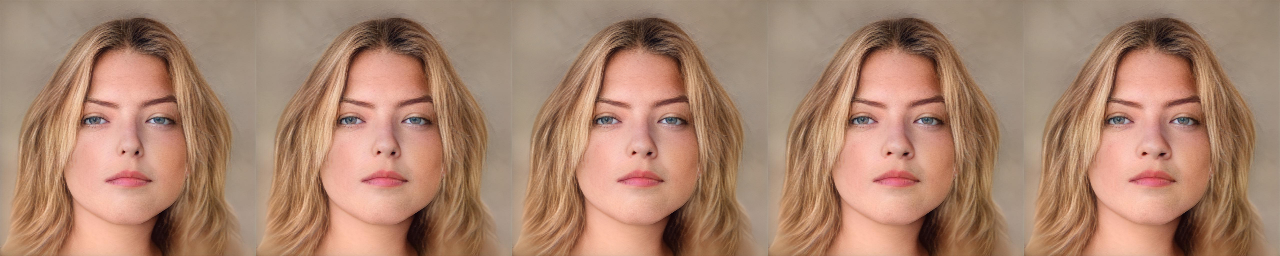}
         \caption{Smaller \(\longleftarrow\) Nose \(\longrightarrow\) Bigger}
         \label{fig:nose}
     \end{subfigure}
     \hfill
     \begin{subfigure}[b]{1\textwidth}
         \centering
         \includegraphics[width=\textwidth]{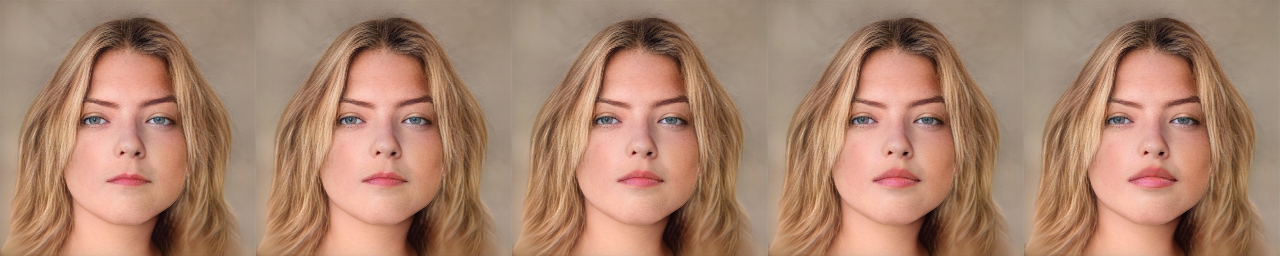}
         \caption{Thickness decrease \(\longleftarrow\) Lips \(\longrightarrow\) Thickness increase}         
         \label{fig:lips}
       \end{subfigure}
     \hfill
     \begin{subfigure}[b]{1\textwidth}
         \centering
         \includegraphics[width=\textwidth]{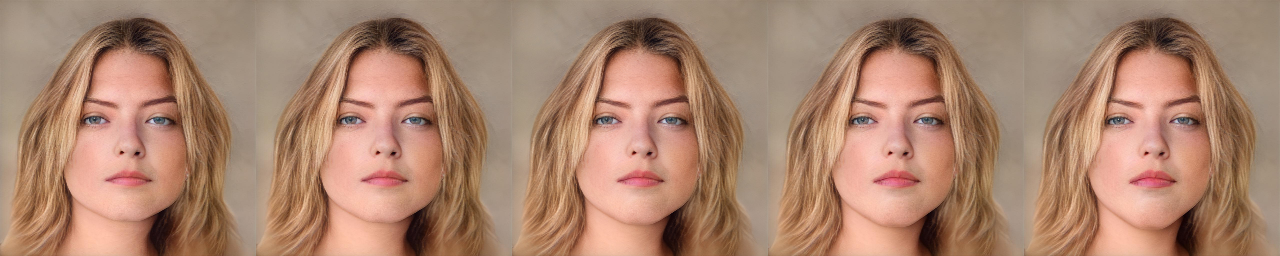}
         \caption{Height decrease \(\longleftarrow\) Chin \(\longrightarrow\) Height increase}         
         \label{fig:chin}
       \end{subfigure}
     \hfill
     \begin{subfigure}[b]{1\textwidth}
         \centering
         \includegraphics[width=\textwidth]{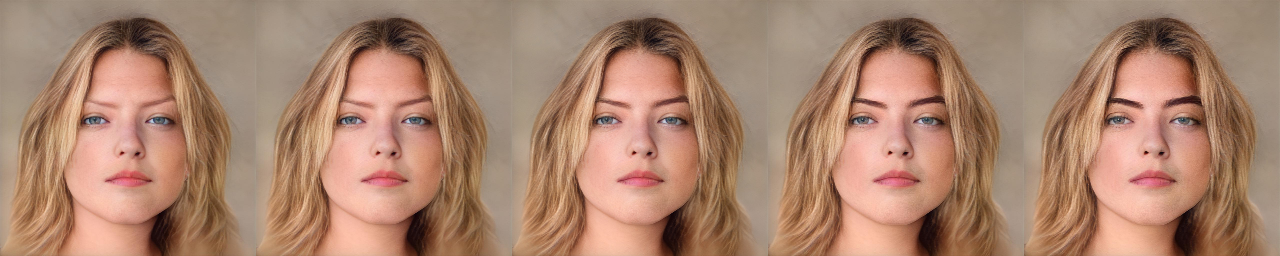}
        \caption{Thickness decrease \(\longleftarrow\) Eyebrow \(\longrightarrow\) Thickness increase }
         \label{fig:eyebrow}
       \end{subfigure}
        \caption{The discovered directions applied to a sample image.}
        \label{fig:sample}
\end{figure*}

Figure \ref{fig:sample} shows the discovered directions applied to a sample image.

\section{Performance Experiments}
\subsection{Methods}
\subsubsection{Dataset}
To perform our experiments, we selected 59 photos from the CelebaA-HQ dataset \cite{karras2018progressive}. The dataset consists of 30 photos of female subjects and 29 male subjects. In selecting the photos, we took into account various parameters such as age and ethnicity to obtain a database that, although small, presented a good variety. Furthermore, to evaluate its applicability in real contexts, we have also selected photos that are not perfectly frontal but with the face slightly turned in one direction. Finally, we have not excluded subjects with accessories such as hats and earrings from the dataset images.

\subsubsection{Photos manipulation}
First, we aligned and then inverted using ReStyle, the 59 photos of our dataset, to obtain the relative latent codes. Therefore, we have regenerated each image through the latent code without alterations (projected version). We, therefore, generated 28 variants of each photo: 4 for each of the 5 directions that we discovered with our method (eye colour, chin size, lip thickness, eyebrows thickness and nose size), 4 for the age direction and 4 for the gender direction. Furthermore, we took advantage of the vectors freely made available in the GitHub repository for the age and gender directions: \url{https://github.com/a312863063/generators-with-stylegan2}.
For each one of the 7 directions analyzed, we used two negative vectors and 2 positive vectors. The sign and magnitude of each variant have been summarized in Table \ref{tab:variations}.

\begin{table}[!htp]\centering
\caption{Magnitude and sign of variation vectors}\label{tab:variations}
\scriptsize
\begin{tabular}{lrrrrr}\toprule
&Variation 1 &Variation 2 &Variation 3 &Variation 4 \\\midrule
Eyes colour &-20 &-10 &10 &20 \\
Chin size &-30 &-15 &15 &30 \\
Lip thickness &-20 &-10 &10 &20 \\
Eyebrows thickness &-40 &-20 &20 &40 \\
Nose Size &-20 &-10 &10 &20 \\
Age &-20 &-10 &10 &20 \\
Gender &-20 &-10 &10 &20 \\
\bottomrule
\end{tabular}
\end{table}

\subsubsection{Face Landmarking}

\begin{figure}[htp]
  \centering
  \includegraphics[width=6cm]{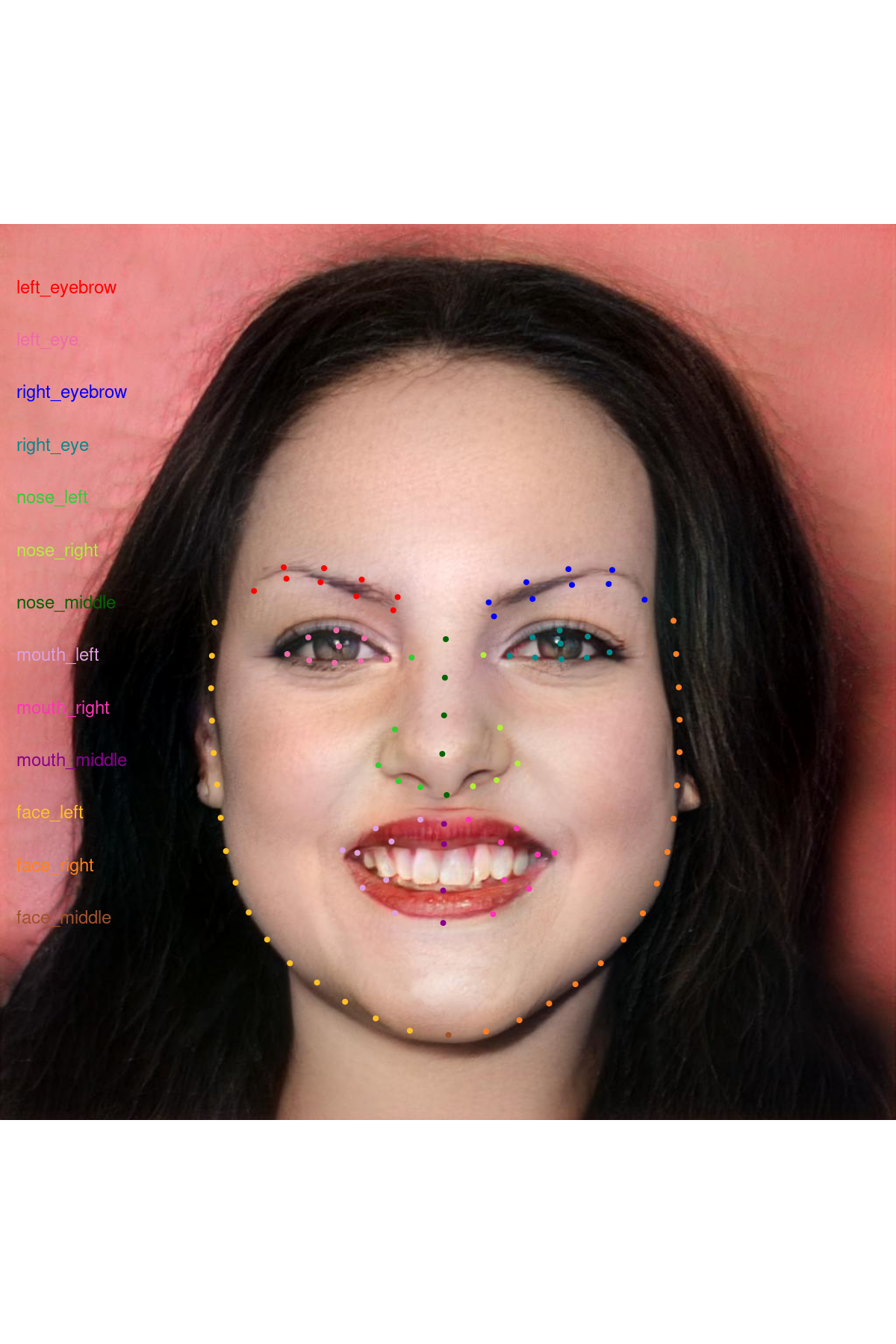}
  \caption{Face++ 106 landmarks}
  \label{fig:face++106}
\end{figure}
\begin{figure}[htp]
  \centering
  \includegraphics[width=6cm]{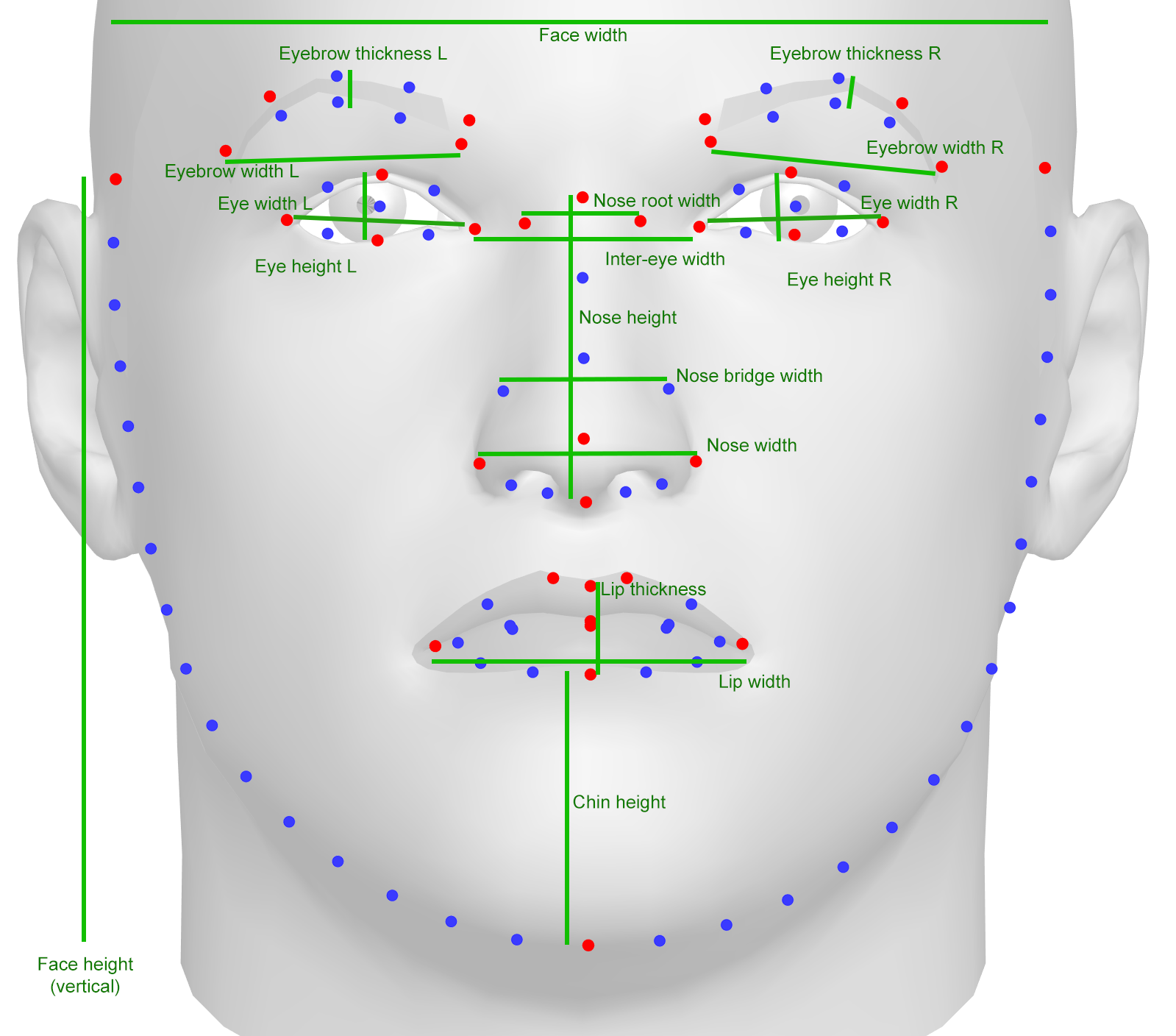}
    \caption{Anatomical landmarks and the distance protocol}
  \label{fig:distances}
\end{figure}

The cloud service platform Face++ (\url{https://www.faceplusplus.com}) was used for landmarking all the images. Landmarks were assigned using Face++’s Detect API. The API also provides attribute values for each image that represent various aspect of image quality (e.g. blurriness, head angle). As the source images were manually curated, all the attributes were within the interval of acceptable values. Face++ placed 106 landmarks on each photograph (Figure \ref{fig:face++106}). We focused on 31 landmarks corresponding mostly to well-defined anatomical landmarks of common usage [\cite{10.1038/ncomms11616},  \cite{doi:10.1126/sciadv.abc6160}, \cite{10.7554/eLife.49898}], and used them to define a set of 18 anatomical measurements (Figure \ref{fig:distances}) which correspond to commonly used dimensions of various facial aspects, and are commonly used in anthropology, such as face width or lip thickness (Table \ref{tab:alignedprojected}). \\

The Dlib (\url{http://dlib.net/} tool and the correlated implementation of the paper ``One Millisecond Face Alignment with an Ensemble of Regression Tree'' \cite{Kazemi_Josephine_2014} have produced the 68 landmarks on the face. Dlib is the tool adopted by StyleGAN authors for automatically face alignment. These are internal landmarks that correspond to key features or contours of the generated face during the construction process, and therefore can be taken as the ground truth. The landmarks implemented in this tool is mostly a subset of the landmarks implemented in Face++. Out of the 31 selected landmarks from Face++, 6 were not available in this protocol, and therefore 4 measurements could not be obtained. \\

\begin{figure}[htp]
  \centering
  \includegraphics[width=6cm]{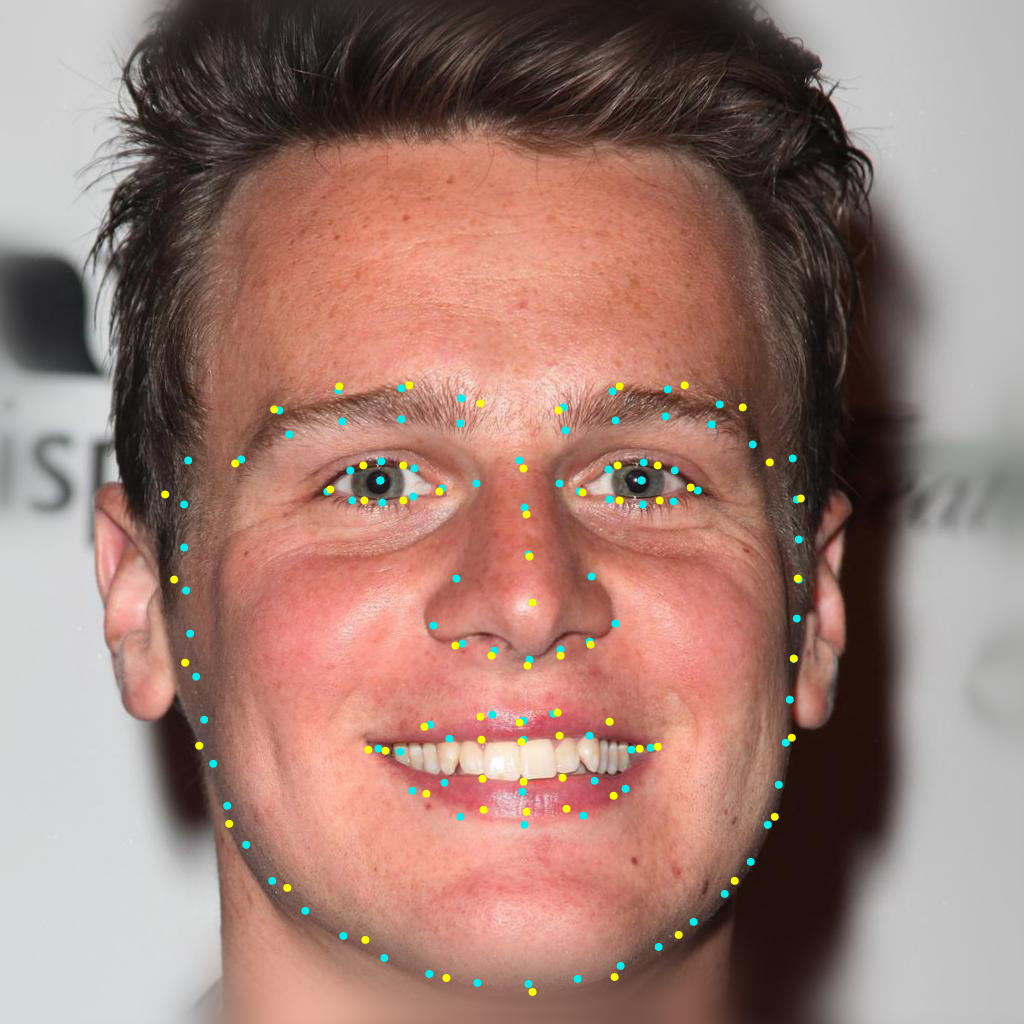}
  \caption{The landmarks used by Face++ in cyan, and by Dlib in yellow}  
  \label{fig:landmarks}
\end{figure}

The landmark coordinates are provided in pixel units, so the measurements are also in pixel units. The generated images were all 1024 X 1024 pixels in resolution, which provided about 25-50 pixel separation between nearby landmarks (Figure \ref{fig:landmarks}). If the generated images were too small in resolution, the landmarks would be too close by, as their coordinates are rounded; if the output resolution was too big then their generation and landmarking would be too time-consuming. \\ 
We checked how accurately the regenerated (projected version) image
matches the aligned image, in terms of landmark placements. For each of the Face++ and
Dlib landmarking system, this was assessed by comparing the distance between the locations
of the same landmark between the two versions. It was measured in pixel units. This was
averaged across all the images to get an average measure of distance.
For comparison, we calculated the average variability of each landmark across all the
projected images, by taking the distance between every pair of images for the same landmark,
and taking the average (Table \ref{tab:distance}). \\
We checked how well the Face++ landmarking performs by
comparing it with the Dlib landmarking results, which is internal to the StyleGAN process
and can be taken as the ground truth. For each of the regenerated (projected version) image,
for each landmark, we calculated the distance between the location of the landmark across the
two protocols. This was averaged across all the regenerated images to get an average measure
of distance, in pixel units (Table \ref{tab:distancecomp}).

\begin{table}[!htp]\centering
\caption{Distance of the landmark across Face++ and Dlib protocols}\label{tab:distancecomp}
\scriptsize
\begin{tabular}{cc}\toprule
\textbf{Landmark (dlib number)} &\textbf{Distance between face++ and dlib} \\\midrule
1 &24.49 \\
9 &9.45 \\
17 &22.4 \\
18 &13.06 \\
22 &24.26 \\
23 &25.71 \\
27 &12.4 \\
28 &8.4 \\
32 &5.05 \\
34 &5.54 \\
36 &4.86 \\
37 &5.16 \\
38 &21.27 \\
40 &7.12 \\
42 &20.99 \\
43 &4.39 \\
44 &22.33 \\
46 &8.66 \\
48 &21.38 \\
49 &7.05 \\
52 &7.5 \\
55 &7.09 \\
58 &7.26 \\
\bottomrule
\end{tabular}
\end{table}

\subsubsection{Analysis of Measurements}
The aligned images have been projected in the StyleGAN2 latent space, using the ReStyle encoder to obtain a latent code and perform the subsequent manipulations. As we said, ReStyle is a learning-based inversion method and not an optimization-based method, so we expect some differences between the aligned image and the projected image. \\
To assess the consistency between the projected image and the original (aligned) image, we calculated correlations of measurements obtained of both versions of each image. This was done for both landmarking protocols. \\
Then, we wanted to assess how perturbing each of the 7 parameters (directions) of the generated faces impact the facial measurements. For example, if we increase the value of the nose width parameter while generation, the nose width measurement of the simulated photos should go up, in other words, this parameter should be positively correlated with nose width. Additionally, if the directions have been established independently of each other, then those parameters should have low correlations with other measurements. For example, changing the nose width or the eye colour parameters should have little impact on the eyebrow thickness measurement.

\onecolumn
\begin{longtable}{p{1.5cm}p{2cm}p{4cm}p{1.5cm}p{2cm}p{4cm}}\toprule
\caption{Distance comparison between the locations of the same landmark between Face++ and Dlib. }\label{tab:distance} \\
\textbf{Face++} &\textbf{} &\textbf{} &\textbf{Dlib} &\textbf{} &\textbf{} \\\midrule
\textbf{Landmark} &\textbf{Variability between images} &\textbf{Change from aligned to projected for an image} &\textbf{Landmark} &\textbf{Variability between images} &\textbf{Change from aligned to projected for an image} \\
1 &46.06 &7.74 &1 &61.4 &9.95 \\
2 &45.7 &7.32 &2 &57.38 &8.91 \\
3 &45.63 &6.98 &3 &57.05 &8.56 \\
4 &45.97 &6.63 &4 &57.54 &8.25 \\
5 &46.73 &6.46 &5 &54.84 &7.59 \\
6 &48.01 &6.3 &6 &49.56 &7.14 \\
7 &49.5 &6.31 &7 &43.48 &7.18 \\
8 &50.81 &6.44 &8 &40.81 &7.25 \\
9 &51.24 &6.55 &9 &41.45 &6.97 \\
10 &50.75 &6.7 &10 &41.36 &6.96 \\
11 &49.29 &6.72 &11 &41.92 &7.01 \\
12 &47.18 &6.81 &12 &44.43 &7.25 \\
13 &44.75 &6.61 &13 &47.28 &7.67 \\
14 &42.22 &6.78 &14 &49.34 &8.01 \\
15 &40.69 &6.91 &15 &49.92 &7.9 \\
16 &41.07 &7.12 &16 &51.92 &8.21 \\
17 &42.1 &7.42 &17 &55.88 &8.63 \\
18 &36.75 &8.33 &18 &30.46 &6.18 \\
19 &36.18 &7.81 &19 &24.47 &4.93 \\
20 &36.03 &7.47 &20 &23.13 &4.23 \\
21 &36.13 &7.24 &21 &25.61 &4.56 \\
22 &36.76 &7.02 &22 &28.12 &5.77 \\
23 &37.85 &6.99 &23 &27.77 &6.58 \\
24 &39.38 &7.19 &24 &25.13 &5.62 \\
25 &40.71 &7.34 &25 &23.09 &5.73 \\
26 &41.4 &7.3 &26 &23.4 &6.4 \\
27 &41.85 &7 &27 &29.01 &7.5 \\
28 &42.04 &6.84 &28 &19.17 &3.19 \\
29 &41.94 &6.7 &29 &26.44 &3.12 \\
30 &41.54 &6.76 &30 &35.55 &3.3 \\
31 &40.84 &6.85 &31 &44.36 &4.22 \\
32 &40.71 &7.09 &32 &28.31 &3.29 \\
33 &41.49 &7.28 &33 &30.32 &2.88 \\
34 &31.23 &8.44 &34 &32.25 &2.94 \\
35 &26.32 &6.27 &35 &30.14 &2.92 \\
36 &23.8 &5.45 &36 &28.45 &3.41 \\
37 &24.23 &5.16 &37 &8.83 &2.98 \\
38 &27.32 &5.96 &38 &8.89 &3.12 \\
39 &27.4 &5.83 &39 &8.68 &3.3 \\
40 &24.53 &4.93 &40 &8.46 &3.35 \\
41 &23.64 &4.77 &41 &7.75 &3.26 \\
42 &26.98 &5.86 &42 &7.42 &3.29 \\
43 &27.79 &6.12 &43 &7.71 &3.18 \\
44 &25.31 &5.12 &44 &7.84 &3.08 \\
45 &24.23 &5.26 &45 &8.66 &3.39 \\
46 &24.91 &5.84 &46 &10.29 &3.52 \\
47 &28.88 &7.51 &47 &8.31 &3.44 \\
48 &26.93 &5.99 &48 &7.91 &3.16 \\
49 &24.36 &4.28 &49 &30.24 &4.33 \\
50 &24.61 &4.45 &50 &28.69 &3.93 \\
51 &25.5 &5.32 &51 &30.49 &4.05 \\
52 &13.93 &3.93 &52 &31.19 &4.04 \\
53 &21.09 &3.59 &53 &30.63 &4.51 \\
54 &30.24 &3.93 &54 &29.85 &4.15 \\
55 &39.99 &4.85 &55 &30.67 &5.13 \\
56 &11.95 &4.42 &56 &31.71 &5.45 \\
57 &24.29 &4 &57 &36.05 &5.5 \\
58 &26.58 &4.59 &58 &36.34 &5.3 \\
59 &28.37 &3.77 &59 &35.5 &5.25 \\
60 &29.9 &3.44 &60 &30.74 &5.21 \\
61 &32.27 &3.43 &61 &29.41 &4.01 \\
62 &11.86 &4.3 &62 &27.87 &3.93 \\
63 &24.25 &4.08 &63 &28.62 &3.87 \\
64 &26.99 &4.65 &64 &28.11 &4.12 \\
65 &28.48 &3.9 &65 &29.61 &4.91 \\
66 &29.9 &3.45 &66 &34.33 &5.09 \\
67 &9.51 &3.53 &67 &34.92 &4.66 \\
68 &8.93 &3.25 &68 &34.18 &4.55 \\
69 &8.51 &3.29 & & & \\
70 &7.66 &3.27 & & & \\
71 &8.83 &3.87 & & & \\
72 &7.36 &3.45 & & & \\
73 &7.14 &3.46 & & & \\
74 &7.78 &3.23 & & & \\
75 &8.21 &3 & & & \\
76 &8.21 &3 & & & \\
77 &8.31 &3.85 & & & \\
78 &7.77 &3.75 & & & \\
79 &8.65 &4.24 & & & \\
80 &9.23 &3.8 & & & \\
81 &10.15 &3.92 & & & \\
82 &8.45 &3.56 & & & \\
83 &7.83 &3.76 & & & \\
84 &7.5 &3.48 & & & \\
85 &9.17 &3.77 & & & \\
86 &9.17 &3.77 & & & \\
87 &30.94 &6.27 & & & \\
88 &29.84 &6.3 & & & \\
89 &26.92 &4.91 & & & \\
90 &27.99 &5.45 & & & \\
91 &26.71 &4.59 & & & \\
92 &29.81 &5.95 & & & \\
93 &30.9 &6.52 & & & \\
94 &29.28 &5.3 & & & \\
95 &27.73 &5.23 & & & \\
96 &28.21 &4.8 & & & \\
97 &30.5 &6.37 & & & \\
98 &28.98 &5.3 & & & \\
99 &30.56 &4.95 & & & \\
100 &30.34 &5.39 & & & \\
101 &33.73 &5.9 & & & \\
102 &29.78 &5.02 & & & \\
103 &30.36 &5.24 & & & \\
104 &33.35 &5.6 & & & \\
105 &33.15 &5.74 & & & \\
106 &35.46 &6.35 & & & \\
\bottomrule
\end{longtable}
\twocolumn

\subsection{Results}
\subsubsection{Analysis of Measurements}
The correlations of measurements between the aligned and projected versions of images were consistently high (Table \ref{tab:alignedprojected}), ranging from \(0.81\) to \(0.98\). This indicated that the projection is not substantially distorting the major aspects of the photographs, and the original measurements are mostly being preserved.
The four parameters that correspond to perturbations of facial measurements – eyebrow, nose, lips, and chin – all had high correlations with their corresponding measurements, according to both landmarking protocols. It indicates that the parameters are working as expected, and they have a roughly linear impact of the target distances (measurements).
These four parameters do not seem to have a huge impact on measurements of other aspects, e.g. changing nose width does not seem to have a big correlation with lip width, even though they are situated close to each other. This validates our working assumption that there is a certain degree of independence between these different facial aspects. Though there seem to be some shared variation, e.g. increasing chin height seems to also increase nose height a bit (Table \ref{tab:paramsmeasures}).

A negative correlation with eye measurements can be noted regarding the blue eyes direction. The negative correlation indicates that, unexpectedly, this direction reduces the size of the eyes. \\

The gender direction, pointing towards masculinity, shows a positive correlation with the size of the face and a negative correlation with the size of the eyes. From this, we can deduce that the gender direction tends to enlarge the face's size and reduce that of the eyes. \\

The age direction, which points towards youth, generally negatively correlates with face size. This is an expected result, given that, particularly on younger subjects, this direction transforms an adult subject into a teenager or a child, and consequently, the size of the face is smaller. \\

\begin{figure}
  \centering
         \includegraphics[width=5cm]{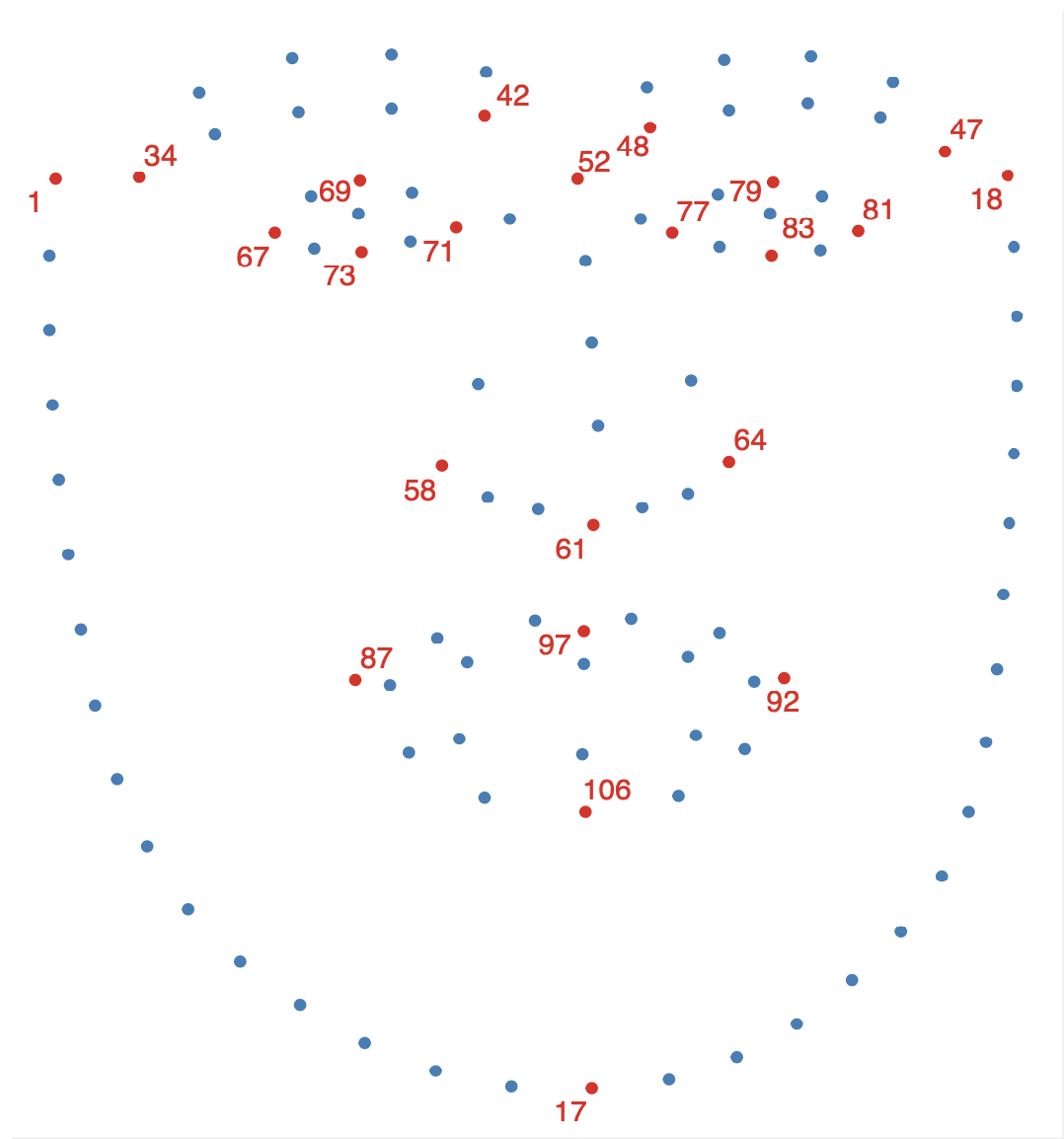}
  \caption{Face++ landmarks. The landmarks that are used in the measurements are in red, and their numbers are shown, whereas the rest of the landmarks are in blue.}
  \label{fig:face++marks}
\end{figure}

\begin{figure}
  \centering
         \includegraphics[width=5cm]{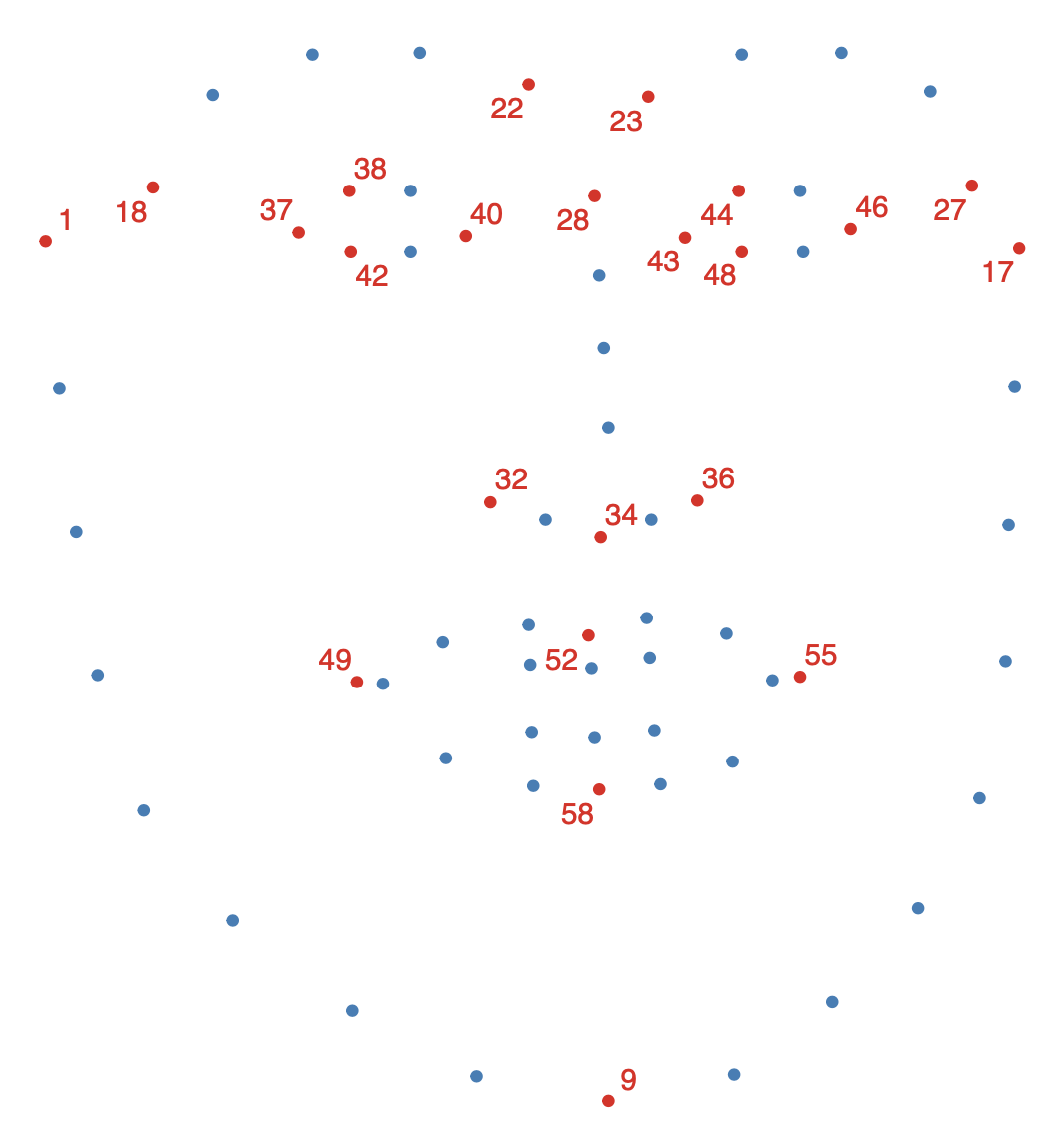}
  \caption{Dlib landmarks. The landmarks that are used in the measurements are in red, and their numbers are shown, whereas the rest of the landmarks are in blue.}
  \label{fig:dlibmarks}
\end{figure}

In the context of 1024×1024 resolution images, the average shift in landmarks between the aligned and regenerated (projected version) images was between 3 to 9 pixels for both protocols, the average being 5.17 in Dlib and 5.44 in Face++. Given the image dimensions, this is around 0.5\%, which seems fairly small.
In comparison, the average variability across projected versions of input images was 31.04 in Dlib and 28.68 in Face++. In other words, the variability incurred by the regeneration (projection) process is much smaller than the original variability in the images, and is therefore acceptable. \\

For assessing the performance of the Face++ landmarking, we compared it to the location of the Dlib landmarks. Some of the landmark locations of the Face++ protocol do not correspond exactly with the Dlib landmarks (see Figure \ref{fig:face++marks} and Figure \ref{fig:dlibmarks}) , so for some of those landmarks we used the position of a nearby semi-landmark instead. The discrepancy ranged from 4 to 26 pixels, the average being 12.86 pixels. Given the image dimensions, this is around 1\% variation, which seems fairly small, and indicates that the Face++ landmarks are acceptably close to the ground truth, at least in our set of projected versions of curated input images. This is also consistent with our observation that the measurements obtained from Face++ landmarks perform similarly to the measurements obtained from Dlib landmarks. This provides implicit evidence of reliability of the Face++ landmarks that have no equivalent in Dlib, such as the ones on the lower eyebrow boundary (useful for measuring eyebrown thickness), or the ones on nose root or nose bridge (useful for measuring breadth of these structures, which have known biological basis and associated genes \cite{10.1038/ncomms11616}.

\begin{table}[!htp]\centering
\caption{Correlation of measurements between aligned and Projected versions of an image}\label{tab:alignedprojected}
\scriptsize
\begin{tabular}{lrrrr}\toprule
\multicolumn{4}{c}{\cellcolor[HTML]{ffffff}\textbf{Correlation of measurements between aligned and Projected versions of an image}} \\\midrule
\textbf{Measurement} &\textbf{Abbreviation} &\textbf{Face++} &\textbf{StyleGAN} \\
\textbf{face width} &\textbf{fw} &\cellcolor[HTML]{f97274}0.94 &\cellcolor[HTML]{f96f71}0.96 \\
\textbf{face height} &\textbf{fh} &\cellcolor[HTML]{f96d6f}0.98 &\cellcolor[HTML]{f97173}0.95 \\
\textbf{eyebrow thickness left} &\textbf{ebtl} &\cellcolor[HTML]{f9797b}0.90 &- \\
\textbf{eyebrow thickness right} &\textbf{ebtr} &\cellcolor[HTML]{f98689}0.80 &- \\
\textbf{eyebrow width left} &\textbf{ebwl} &\cellcolor[HTML]{f97173}0.95 &\cellcolor[HTML]{f96e70}0.97 \\
\textbf{eyebrow width right} &\textbf{ebwr} &\cellcolor[HTML]{f97375}0.93 &\cellcolor[HTML]{f97173}0.95 \\
\textbf{eye width left} &\textbf{ewl} &\cellcolor[HTML]{f97c7e}0.88 &\cellcolor[HTML]{f97779}0.91 \\
\textbf{eye width right} &\textbf{ewr} &\cellcolor[HTML]{f97e80}0.86 &\cellcolor[HTML]{f97678}0.92 \\
\textbf{eye height left} &\textbf{ehl} &\cellcolor[HTML]{f97e80}0.86 &\cellcolor[HTML]{f97173}0.95 \\
\textbf{eye height right} &\textbf{ehr} &\cellcolor[HTML]{f98587}0.81 &\cellcolor[HTML]{f97476}0.93 \\
\textbf{inter-eye width} &\textbf{iew} &\cellcolor[HTML]{f98285}0.83 &\cellcolor[HTML]{f97678}0.92 \\
\textbf{nose root width} &\textbf{nrw} &\cellcolor[HTML]{f98385}0.83 &- \\
\textbf{nose bridge width} &\textbf{nbw} &\cellcolor[HTML]{f9777a}0.90 &- \\
\textbf{nose width} &\textbf{nw} &\cellcolor[HTML]{f97274}0.94 &\cellcolor[HTML]{f96f71}0.96 \\
\textbf{nose height} &\textbf{nh} &\cellcolor[HTML]{f96c6e}0.98 &\cellcolor[HTML]{f96e70}0.97 \\
\textbf{lip thickness} &\textbf{lt} &\cellcolor[HTML]{f96e70}0.97 &\cellcolor[HTML]{f96e70}0.97 \\
\textbf{lip width} &\textbf{lw} &\cellcolor[HTML]{f96e70}0.97 &\cellcolor[HTML]{f96c6e}0.98 \\
\textbf{chin height} &\textbf{ch} &\cellcolor[HTML]{f96f71}0.96 &\cellcolor[HTML]{f96d6f}0.98 \\
\bottomrule
\end{tabular}
\end{table}

\begin{table*}[!htp]\centering
\caption{Correlation between parameters and measurements}\label{tab:paramsmeasures}
\scriptsize
\begin{tabular}{lrrrrrrrrrrrrrrr}\toprule
\multicolumn{15}{c}{\textbf{Correlation between parameters and measurements}} \\\cmidrule{1-15}
\textbf{} &\multicolumn{7}{c}{\cellcolor[HTML]{3ee8e9}\textbf{Using Face++ landmarks}} &\multicolumn{7}{c}{\cellcolor[HTML]{e0ec25}\textbf{Using StyleGAN internal landmarks}} \\\cmidrule{2-15}
\textbf{Measurement} &\textbf{age} &\textbf{gender} &\textbf{blueeyes} &\textbf{eyebrow} &\textbf{nose} &\textbf{lips} &\textbf{chin} &\textbf{age} &\textbf{gender} &\textbf{blueeyes} &\textbf{eyebrow} &\textbf{nose} &\textbf{lips} &\textbf{chin} \\\midrule
\textbf{face width} &\cellcolor[HTML]{aac2e2}-0.50 &\cellcolor[HTML]{fab0b2}0.52 &\cellcolor[HTML]{dbe5f3}-0.20 &\cellcolor[HTML]{fbced1}0.31 &\cellcolor[HTML]{eaeff8}-0.11 &\cellcolor[HTML]{e3eaf6}-0.15 &\cellcolor[HTML]{fcdbde}0.23 &\cellcolor[HTML]{c1d3ea}-0.36 &\cellcolor[HTML]{fa9c9e}0.66 &\cellcolor[HTML]{e3eaf6}-0.15 &\cellcolor[HTML]{fbbec1}0.42 &\cellcolor[HTML]{e5ebf6}-0.14 &\cellcolor[HTML]{d2def0}-0.26 &\cellcolor[HTML]{fce3e6}0.17 \\
\textbf{face height} &\cellcolor[HTML]{6995cb}-0.90 &\cellcolor[HTML]{fcdfe2}0.20 &\cellcolor[HTML]{fbfbfe}-0.00 &\cellcolor[HTML]{fceaed}0.12 &\cellcolor[HTML]{fcfafd}0.02 &\cellcolor[HTML]{f8f9fd}-0.02 &\cellcolor[HTML]{fbd6d8}0.26 &\cellcolor[HTML]{8aacd7}-0.70 &\cellcolor[HTML]{fa9d9f}0.65 &\cellcolor[HTML]{f5f7fc}-0.04 &\cellcolor[HTML]{fbc1c4}0.40 &\cellcolor[HTML]{fbcacd}0.34 &\cellcolor[HTML]{e2eaf6}-0.16 &\cellcolor[HTML]{f97f81}0.86 \\
\textbf{eyebrow thickness left} &\cellcolor[HTML]{c6d6ec}-0.33 &\cellcolor[HTML]{faa7aa}0.58 &\cellcolor[HTML]{f6f8fd}-0.03 &\cellcolor[HTML]{fbbabc}0.45 &\cellcolor[HTML]{fcf4f7}0.06 &\cellcolor[HTML]{fbfbfe}-0.00 &\cellcolor[HTML]{fcdfe2}0.20 &- &- &- &- &- &- &- \\
\textbf{eyebrow thickness right} &\cellcolor[HTML]{f6f8fd}-0.03 &\cellcolor[HTML]{fbb5b8}0.49 &\cellcolor[HTML]{fafafe}-0.01 &\cellcolor[HTML]{faadaf}0.54 &\cellcolor[HTML]{fce2e5}0.18 &\cellcolor[HTML]{f0f3fa}-0.07 &\cellcolor[HTML]{fcf3f6}0.06 &- &- &- &- &- &- &- \\
\textbf{eyebrow width left} &\cellcolor[HTML]{fce5e7}0.16 &\cellcolor[HTML]{fcdddf}0.22 &\cellcolor[HTML]{dee7f4}-0.18 &\cellcolor[HTML]{fa989a}0.68 &\cellcolor[HTML]{eef2fa}-0.08 &\cellcolor[HTML]{d5e0f1}-0.24 &\cellcolor[HTML]{fcf0f3}0.09 &\cellcolor[HTML]{cedbee}-0.28 &\cellcolor[HTML]{fa9699}0.70 &\cellcolor[HTML]{e3eaf6}-0.15 &\cellcolor[HTML]{fab0b2}0.52 &\cellcolor[HTML]{e7edf7}-0.13 &\cellcolor[HTML]{b7cce7}-0.42 &\cellcolor[HTML]{fcfcff}0.00 \\
\textbf{eyebrow width right} &\cellcolor[HTML]{fcdadd}0.23 &\cellcolor[HTML]{fcdde0}0.22 &\cellcolor[HTML]{f8f9fd}-0.02 &\cellcolor[HTML]{faadaf}0.54 &\cellcolor[HTML]{fce9ec}0.13 &\cellcolor[HTML]{d5e1f1}-0.23 &\cellcolor[HTML]{fcf5f8}0.05 &\cellcolor[HTML]{fbfbfe}-0.00 &\cellcolor[HTML]{fbb6b8}0.48 &\cellcolor[HTML]{cbdaee}-0.30 &\cellcolor[HTML]{faa3a5}0.61 &\cellcolor[HTML]{fcecef}0.11 &\cellcolor[HTML]{cddbee}-0.28 &\cellcolor[HTML]{e9eff8}-0.11 \\
\textbf{eye width left} &\cellcolor[HTML]{d9e3f2}-0.21 &\cellcolor[HTML]{cfdcef}-0.28 &\cellcolor[HTML]{b8cce7}-0.42 &\cellcolor[HTML]{fce4e6}0.17 &\cellcolor[HTML]{f4f6fc}-0.04 &\cellcolor[HTML]{fcf5f8}0.05 &\cellcolor[HTML]{f9fafe}-0.01 &\cellcolor[HTML]{fcf8fb}0.03 &\cellcolor[HTML]{c2d3ea}-0.36 &\cellcolor[HTML]{c6d6ec}-0.33 &\cellcolor[HTML]{fcfbfe}0.01 &\cellcolor[HTML]{fcfcff}0.00 &\cellcolor[HTML]{f7f9fd}-0.03 &\cellcolor[HTML]{f6f8fd}-0.03 \\
\textbf{eye width right} &\cellcolor[HTML]{fcdcdf}0.22 &\cellcolor[HTML]{b7cbe6}-0.42 &\cellcolor[HTML]{a3bddf}-0.55 &\cellcolor[HTML]{fcdadd}0.23 &\cellcolor[HTML]{eef2fa}-0.08 &\cellcolor[HTML]{e1e9f5}-0.17 &\cellcolor[HTML]{d9e3f2}-0.21 &\cellcolor[HTML]{fceaed}0.13 &\cellcolor[HTML]{bfd1e9}-0.38 &\cellcolor[HTML]{c4d5eb}-0.34 &\cellcolor[HTML]{fceef1}0.10 &\cellcolor[HTML]{fcf8fb}0.03 &\cellcolor[HTML]{fcfcff}0.00 &\cellcolor[HTML]{f7f8fd}-0.03 \\
\textbf{eye height left} &\cellcolor[HTML]{e3eaf6}-0.15 &\cellcolor[HTML]{88aad6}-0.71 &\cellcolor[HTML]{90b0d9}-0.66 &\cellcolor[HTML]{fceaed}0.12 &\cellcolor[HTML]{fcfafd}0.02 &\cellcolor[HTML]{e6ecf7}-0.13 &\cellcolor[HTML]{e9eff8}-0.11 &\cellcolor[HTML]{d4dff0}-0.25 &\cellcolor[HTML]{84a7d4}-0.74 &\cellcolor[HTML]{b5cae6}-0.43 &\cellcolor[HTML]{fcf9fc}0.03 &\cellcolor[HTML]{e3eaf6}-0.15 &\cellcolor[HTML]{e6edf7}-0.13 &\cellcolor[HTML]{f1f4fb}-0.06 \\
\textbf{eye height right} &\cellcolor[HTML]{c7d6ec}-0.33 &\cellcolor[HTML]{92b1d9}-0.65 &\cellcolor[HTML]{96b4db}-0.63 &\cellcolor[HTML]{fcdadd}0.23 &\cellcolor[HTML]{fcf8fb}0.03 &\cellcolor[HTML]{e1e9f5}-0.16 &\cellcolor[HTML]{e4ebf6}-0.14 &\cellcolor[HTML]{e3eaf6}-0.15 &\cellcolor[HTML]{8bacd7}-0.69 &\cellcolor[HTML]{aac2e2}-0.50 &\cellcolor[HTML]{fcf4f7}0.06 &\cellcolor[HTML]{e2eaf6}-0.15 &\cellcolor[HTML]{e7edf7}-0.13 &\cellcolor[HTML]{eff2fa}-0.08 \\
\textbf{inter-eye width} &\cellcolor[HTML]{f1f4fb}-0.06 &\cellcolor[HTML]{fceaed}0.12 &\cellcolor[HTML]{faadaf}0.54 &\cellcolor[HTML]{fcedef}0.11 &\cellcolor[HTML]{fcf1f4}0.08 &\cellcolor[HTML]{dee7f4}-0.18 &\cellcolor[HTML]{fbd8da}0.25 &\cellcolor[HTML]{fcebee}0.12 &\cellcolor[HTML]{fce9ec}0.13 &\cellcolor[HTML]{fbcfd2}0.31 &\cellcolor[HTML]{fbd0d3}0.30 &\cellcolor[HTML]{fcdee1}0.21 &\cellcolor[HTML]{dde6f4}-0.19 &\cellcolor[HTML]{fce0e3}0.19 \\
\textbf{nose root width} &\cellcolor[HTML]{fbcbce}0.34 &\cellcolor[HTML]{fbd1d4}0.30 &\cellcolor[HTML]{f3f5fb}-0.05 &\cellcolor[HTML]{fcf7fa}0.03 &\cellcolor[HTML]{fa8e91}0.75 &\cellcolor[HTML]{bfd1e9}-0.38 &\cellcolor[HTML]{fce3e6}0.17 &- &- &- &- &- &- &- \\
\textbf{nose bridge width} &\cellcolor[HTML]{fcf6f9}0.04 &\cellcolor[HTML]{fbc3c6}0.39 &\cellcolor[HTML]{f2f5fb}-0.06 &\cellcolor[HTML]{fafafe}-0.01 &\cellcolor[HTML]{f9797b}0.90 &\cellcolor[HTML]{cddaee}-0.29 &\cellcolor[HTML]{fceef1}0.10 &- &- &- &- &- &- &- \\
\textbf{nose width} &\cellcolor[HTML]{b5cae6}-0.43 &\cellcolor[HTML]{fbd0d3}0.30 &\cellcolor[HTML]{ebf0f9}-0.10 &\cellcolor[HTML]{fce3e6}0.17 &\cellcolor[HTML]{f97779}0.91 &\cellcolor[HTML]{e1e9f5}-0.16 &\cellcolor[HTML]{fcfafd}0.01 &\cellcolor[HTML]{7ea3d2}-0.78 &\cellcolor[HTML]{fbc9cb}0.35 &\cellcolor[HTML]{fcf1f4}0.08 &\cellcolor[HTML]{f7f8fd}-0.03 &\cellcolor[HTML]{f97375}0.93 &\cellcolor[HTML]{e8edf7}-0.12 &\cellcolor[HTML]{f0f3fa}-0.07 \\
\textbf{nose height} &\cellcolor[HTML]{749ccf}-0.83 &\cellcolor[HTML]{fcf0f3}0.09 &\cellcolor[HTML]{fcdbde}0.23 &\cellcolor[HTML]{fbc0c2}0.41 &\cellcolor[HTML]{fa9d9f}0.65 &\cellcolor[HTML]{f3f5fb}-0.05 &\cellcolor[HTML]{fab1b4}0.51 &\cellcolor[HTML]{7099cd}-0.86 &\cellcolor[HTML]{dde6f4}-0.19 &\cellcolor[HTML]{fbd2d4}0.29 &\cellcolor[HTML]{fbcccf}0.33 &\cellcolor[HTML]{faa6a8}0.59 &\cellcolor[HTML]{b0c6e4}-0.47 &\cellcolor[HTML]{fbbabc}0.45 \\
\textbf{lip thickness} &\cellcolor[HTML]{fa9c9f}0.65 &\cellcolor[HTML]{afc6e4}-0.47 &\cellcolor[HTML]{f6f7fc}-0.04 &\cellcolor[HTML]{fcf2f5}0.07 &\cellcolor[HTML]{fcf4f6}0.06 &\cellcolor[HTML]{f98082}0.85 &\cellcolor[HTML]{faaaad}0.56 &\cellcolor[HTML]{f97d7f}0.87 &\cellcolor[HTML]{94b2da}-0.64 &\cellcolor[HTML]{f2f5fb}-0.06 &\cellcolor[HTML]{fcf9fc}0.02 &\cellcolor[HTML]{fceff2}0.09 &\cellcolor[HTML]{f9797b}0.90 &\cellcolor[HTML]{fa9699}0.70 \\
\textbf{lip width} &\cellcolor[HTML]{81a6d4}-0.75 &\cellcolor[HTML]{c3d4eb}-0.35 &\cellcolor[HTML]{ccdaee}-0.29 &\cellcolor[HTML]{f6f7fc}-0.04 &\cellcolor[HTML]{fcecee}0.11 &\cellcolor[HTML]{fa9ea0}0.65 &\cellcolor[HTML]{fce3e6}0.17 &\cellcolor[HTML]{6e98cd}-0.87 &\cellcolor[HTML]{b8cce7}-0.42 &\cellcolor[HTML]{dbe5f3}-0.20 &\cellcolor[HTML]{fce2e5}0.18 &\cellcolor[HTML]{fbcdd0}0.32 &\cellcolor[HTML]{faa0a2}0.63 &\cellcolor[HTML]{fce2e5}0.18 \\
\textbf{chin height} &\cellcolor[HTML]{6a95cb}-0.90 &\cellcolor[HTML]{fa8f91}0.75 &\cellcolor[HTML]{fbfbfe}-0.00 &\cellcolor[HTML]{fbced1}0.32 &\cellcolor[HTML]{eaeff8}-0.11 &\cellcolor[HTML]{a3bddf}-0.55 &\cellcolor[HTML]{fa8f91}0.74 &\cellcolor[HTML]{618ec8}-0.96 &\cellcolor[HTML]{f97e80}0.86 &\cellcolor[HTML]{f2f5fb}-0.06 &\cellcolor[HTML]{fbbcbf}0.44 &\cellcolor[HTML]{fcfcff}0.01 &\cellcolor[HTML]{98b6dc}-0.61 &\cellcolor[HTML]{f9878a}0.80 \\
\bottomrule
\end{tabular}
\end{table*}

\section{Conclusion and future work}
We believe that the method has ample room for improvement. The eye colour direction did not prove solid enough, as it changed not only the eye colour but also the eyes size. In a future version, we hope to solve this problem through more accurate photo editing in the models used to calculate the directions. In future work, we will also explore the method applied to the newer StyleGAN3 implementation \cite{karras2021aliasfree}. \\
The discovery of new directions to have accurate control of the somatic characteristics of a face remains an open field and will undoubtedly be the subject of future developments. In particular, we think that better results can also be obtained by adopting inversion methods other than ReStyle, such as the one proposed by Yangyang Xu \textit{et al} in the paper ``From Continuity to Editability: Inverting GANs With Consecutive Images'' \cite{Xu_2021_ICCV}. If, on the one hand, ReStyle has proved excellent in generalizing semantic changes and, therefore, in the calculation of directional vectors, on the other side, it has shown limits in reversing photos of subjects in natural conditions. In particular, we noticed several problems in the inversion of subjects with accessories, such as earrings or hats. \\
However, we believe that beyond the above problems, the method remains valid and once refined, it can find different practical applications, ranging from forensic investigation to cosmetic surgery.

\section{Code availability}
The code developed here is available at: \\\url{https://github.com/agiardina/stylegan2-directions}

\section{Acknowledgements}
We thank Andres Ruiz-Linares, the principal investigator of CANDELA, for his support. We thank Qing Li for her exploration of Face++.

\bibliography{refs}{}
\bibliographystyle{plain}
\end{document}